%% file: ieee_trans.tex
\documentclass[lettersize,journal]{IEEEtran}
\usepackage{amsmath,amsfonts}
\usepackage{algorithmic}
\usepackage{algorithm}
\usepackage{array}
\usepackage[caption=false,font=normalsize,labelfont=sf,textfont=sf]{subfig}
\usepackage{textcomp}
\usepackage{stfloats}
\usepackage{url}
\usepackage{verbatim}
\usepackage{graphicx}
\usepackage{cite}
\usepackage{xspace}
\usepackage{bm}
\usepackage{amsmath}
\usepackage{CJKutf8}
\usepackage{multirow}
\usepackage{bbding}
\usepackage[figuresright]{rotating}
\usepackage{makecell}
\usepackage{colortbl}
\usepackage{booktabs}
\usepackage{xcolor}

\definecolor{citecolor}{HTML}{000080}
\definecolor{linkcolor}{HTML}{c0392b}

\newcommand{\method}{NFLAT\xspace}

\begin{document}
\begin{CJK}{UTF8}{gbsn}

\title{NFLAT: Non-Flat-Lattice Transformer\\ for Chinese Named Entity Recognition}

\author{
    Shuang Wu,
    Xiaoning Song$^*$,~\IEEEmembership{Member,~IEEE},
	Zhen-Hua~Feng$^*$,~\IEEEmembership{Senior Member,~IEEE},
	and~Xiao-Jun~Wu,~\IEEEmembership{Member,~IEEE}
\thanks{$*$ Corresponding authors}
\thanks{This work was supported by the Major Project of National Social Science Foundation of China (No. 21\&ZD166), the National Natural Science Foundation of China (61876072, 61902153) and the Natural Science Foundation of Jiangsu Province (No. BK20221535).}
\thanks{Shuang Wu, Xiaoning Song and Xiao-Jun Wu are with the School of Artificial Intelligence and Computer Science, Jiangnan University, Wuxi 214122, China. (email: shuangwu@stu.jiangnan.edu.cn, x.song@jiangnan.edu.cn, wu\_xiaojun@jiangnan.edu.cn)}
\thanks{Zhenhua Feng is with the School of Computer Science and Electronic Engineering, University of Surrey, Guildford, GU2 7XH, UK. (e-mail: z.feng@surrey.ac.uk)}
}

\markboth{Journal of \LaTeX\ Class Files,~Vol.~14, No.~8, August~2021}%
{Shell \MakeLowercase{\textit{et al.}}: A Sample Article Using IEEEtran.cls for IEEE Journals}

\IEEEpubid{0000--0000/00\$00.00~\copyright~2021 IEEE}

\maketitle

\begin{abstract}
Recently, Flat-LAttice Transformer (FLAT) has achieved great success in Chinese Named Entity Recognition (NER). FLAT performs lexical enhancement by constructing flat lattices, which mitigates the difficulties posed by blurred word boundaries and the lack of word semantics. In FLAT, the positions of starting and ending characters are used to connect a matching word. However, this method is likely to match more words when dealing with long texts, resulting in long input sequences. Therefore, it significantly increases the memory and computational costs of the self-attention module. To deal with this issue, we advocate a novel lexical enhancement method, InterFormer, that effectively reduces the amount of computational and memory costs by constructing non-flat lattices. Furthermore, with InterFormer as the backbone, we implement \method for Chinese NER. \method decouples lexicon fusion and context feature encoding. Compared with FLAT, it reduces unnecessary attention calculations in ``word-character'' and ``word-word''. This reduces the memory usage by about 50\% and can use more extensive lexicons or higher batches for network training. The experimental results obtained on several well-known benchmarks demonstrate the superiority of the proposed method over the state-of-the-art hybrid (character-word) models. The source code of the proposed method is publicly available at \url{https://github.com/CoderMusou/NFLAT4CNER}.
\end{abstract}


\section{Introduction}
\IEEEPARstart{N}{amed} Entity Recognition (NER) is usually handled as a sequence tagging task, which plays an essential role in Natural Language Processing (NLP).
NER often extracts valuable information from unstructured text, which can be used for many other high-level tasks, such as information retrieval~\cite{khalid-etal-2008-impact}, knowledge graph~\cite{riedel-etal-2013-relation}, question answering~\cite{diefenbach-etal-2018-core}, public opinion analysis~\cite{wang-etal-2016-hybrid}, biomedicine~\cite{settles-2004-biomedical}, recommendation system~\cite{karatay-karagoz-2015-user} and many others.

\begin{figure}[!t]
    \centering
    \subfloat[]{\includegraphics[width=3in]{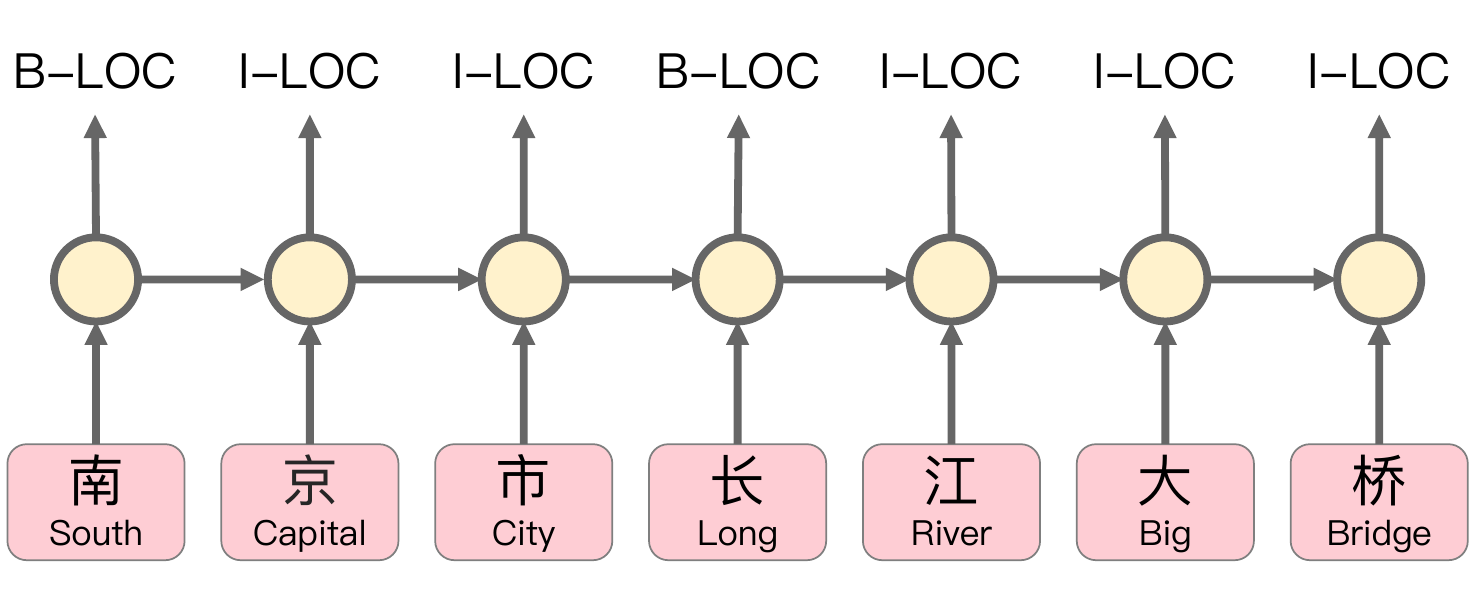}
    \label{fig:char-model}}
    
    \subfloat[]{\includegraphics[width=3in]{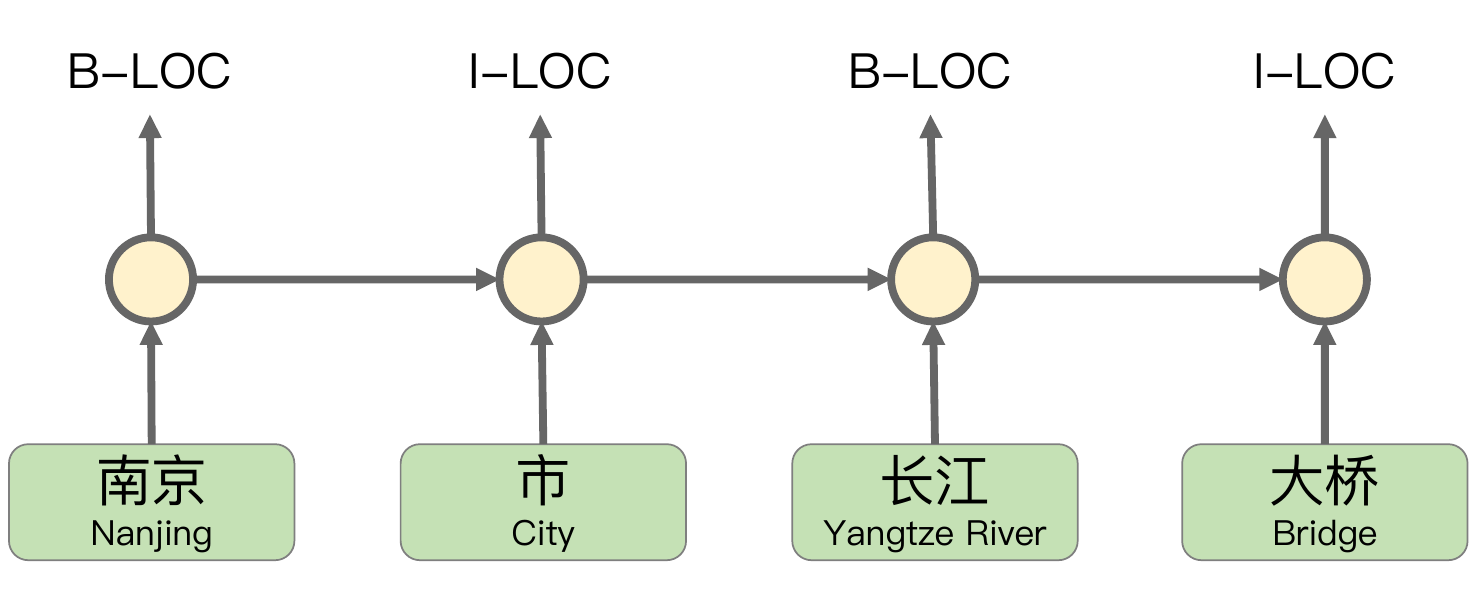}
    \label{fig:word-model}}
    \caption{Chinese NER models with different granularities. The red is character, and the green is word. (a) Character-based model. (b) Word-based model.}
    \label{fig:chinese-ner}
\end{figure}

Compared with English~\cite{lample-etal-2016-neural,ma-hovy-2016-end,chiu-nichols-2016-named,jie-lu-2019-dependency,liu-etal-2019-towards,sun-etal-2020-learning}, Chinese NER is more challenging. 
First, Chinese word boundaries are blurred, and there is no separator, such as space, to clarify word boundaries.
If a character-level model (Fig.~\ref{fig:char-model}) is used for Chinese NER, it will have the problem of missing word semantics and boundary information.
On the other hand, if we use word-level models (Fig.~\ref{fig:word-model}), wrong word segmentation will also degrade the performance. 
In addition, there are more complex properties in Chinese, such as complex combinations, entity nesting, indefinite length~\cite{dong-etal-2016-character}, and network neologisms.
Moreover, Chinese does not have case-sensitive and root-affix properties and lacks the expression of much semantic information.
Therefore, in recent years, the mainstream Chinese NER methods have focused on the use of external data, such as lexicon information~\cite{zhang-yang-2018-chinese}, glyph information~\cite{meng-etal-2019-glyce,wu-etal-2021-mect}, syntactic information~\cite{nie-etal-2020-improving}, and semantic information~\cite{nie-etal-2020-named}, for performance boosting.

Flat-LAttice Transformer (FLAT)~\cite{li-etal-2020-flat} is a very popular lexical enhancement method that effectively extracts entity boundaries and rich word semantics.
However, FLAT increases the computational and memory costs significantly.
Also, it is very difficult to use large-scale lexicons in FLAT.
To mitigate these issues, we propose a novel and efficient lexical enhancement method, \method, that uses a non-flat-lattice architecture.
\IEEEpubidadjcol
The main contributions include:
\begin{itemize}
    \item We propose a new InterFormer network with a non-flat-lattice architecture, which jointly models character and word sequences at different lengths.

    \item With InterFormer as the backbone, we further develop a non-flat lexical enhancement method, namely \method, for Chinese NER.

    \item To achieve further performance boosting, we use an extra tag to help the model automatically suppress noisy information and discover correct words.

    \item We evaluate \method on several well-known benchmarks with different lexicons and the results demonstrate the superiority of our method over the state-of-the-art approaches.
\end{itemize}

\section{Related Work}
\paragraph{Deep Learning in Chinese NER}
Wu \textit{et al.}\cite{wu-etal-2015-named} firstly applied deep networks to extract critical clinical information from Chinese documents with NER.
Peng and Dredze\cite{peng-dredze-2016-improving} proposed a multi-task learning method based on LSTM-CRF, which improves the performance of NER by jointly training it with the word segmentation task.
Inspired by the character-level features in English NER, Lample \textit{et al.}\cite{lample-etal-2016-neural} and Dong \textit{et al.}\cite{dong-etal-2016-character} designed a BiLSTM-CRF model that introduces features containing both characters and radicals.
He and Sun\cite{he-sun-2017-f, he-sun-2017-unified} tried to improve the NER performance in Chinese social media using cross-domain, semi-supervised data and rich embedding features based on BiLSTM-MMNN.
After that, Zhang and Yang\cite{zhang-yang-2018-chinese} proposed a character-word hybrid model, Lattice-LSTM, for Chinese NER.
This method augments the character-level model with lexicon information, avoiding potential segmentation errors in word-level models and missing boundaries in character-level models.
It demonstrates the merits of using lexicon information in Chinese NER.

\paragraph{Lexical Enhancement Methods}
Besides Lattice-LSTM, existing lexical enhancement methods also use many other network architectures, such as CNN in LR-CNN~\cite{gui-etal-2019-cnn}, graph networks in LGN~\cite{gui-etal-2019-lexicon}, CGN~\cite{sui-etal-2019-leverage} and the Neural Multi-digraph Model (NMDM)~\cite{ding-etal-2019-neural}.
More recently, the popular Transformer model has also been used for lexical enhancement, such as PLTE~\cite{mengge-etal-2020-porous} and FLAT~\cite{li-etal-2020-flat}.
Note that some methods, \textit{e.g.}, WC-LSTM~\cite{liu-etal-2019-encoding} and SoftLexicon~\cite{ma-etal-2020-simplify}, fuse lexicon information at the embedding layer.

FLAT is the most relevant study to the proposed method. FLAT significantly improves the performance of Chinese NER by introducing lexicon information via a flat lattice with two positional encodings.
However, this method increases the input sequence length by more than 40\% on average, resulting in several practical issues. First, using longer sequences significantly increases the memory and computational costs in self-attention.
Second, FLAT is unable to use a larger or more comprehensive lexicon.
Last, it unnecessarily computes attention scores for ``word-character'' and ``word-word''.
To address these issues, we propose a novel lexical enhancement method with a more flexible structure and superior performance over the above methods.

\section{Background}
FLAT achieves relatively high performance from scratch and further performance boosting when used with BERT.
Since the underlying structure of FLAT is based on Transformer~\cite{vaswani-etal-2017-attention}, FLAT can extract robust features with high efficiency.
The self-attention mechanism is the key of Transformer, which establishes a connection between each pair of tokens of the input.
Transformer is different from RNN~\cite{elman-1990-finding}, LSTM~\cite{hochreiter-schmidhuber-1997-long}, GRU~\cite{chung-etal-2014-empirical} and other recurrent neural networks, in which the input at each moment needs to depend on the output of the previous moment.
Additionally, Transformer captures long-range dependencies of deep feature maps, resulting in better performance than CNNs and RNNs.

\begin{figure}
    \centering
    \subfloat[]{\includegraphics[width=3.5in]{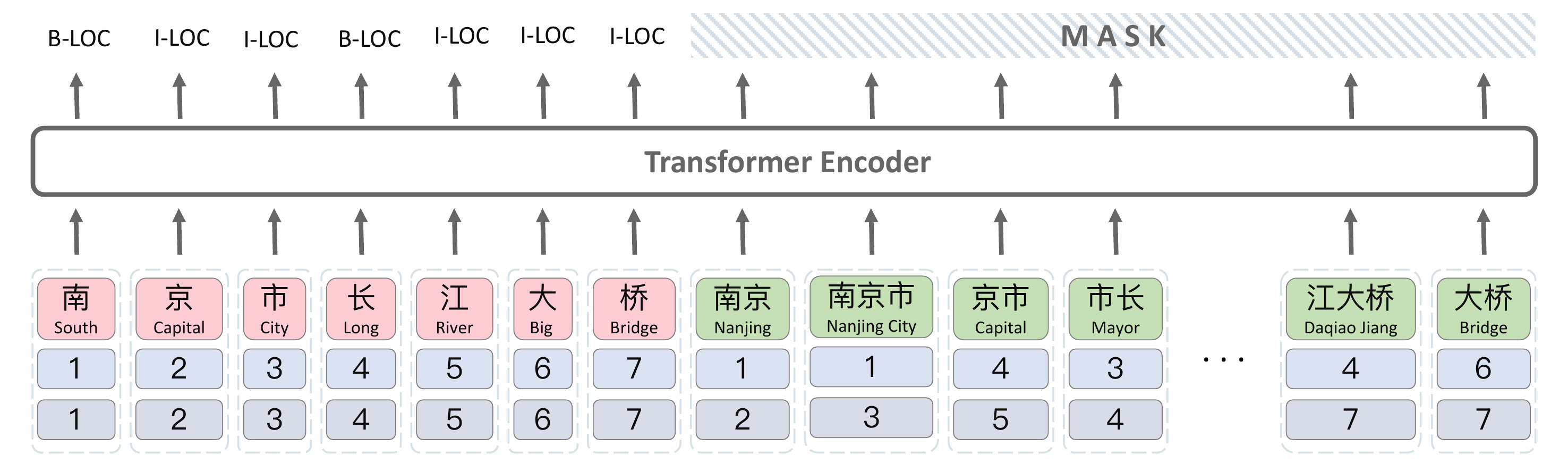}
    \label{fig:flat}}\qquad
    \subfloat[]{\includegraphics[width=3.5in]{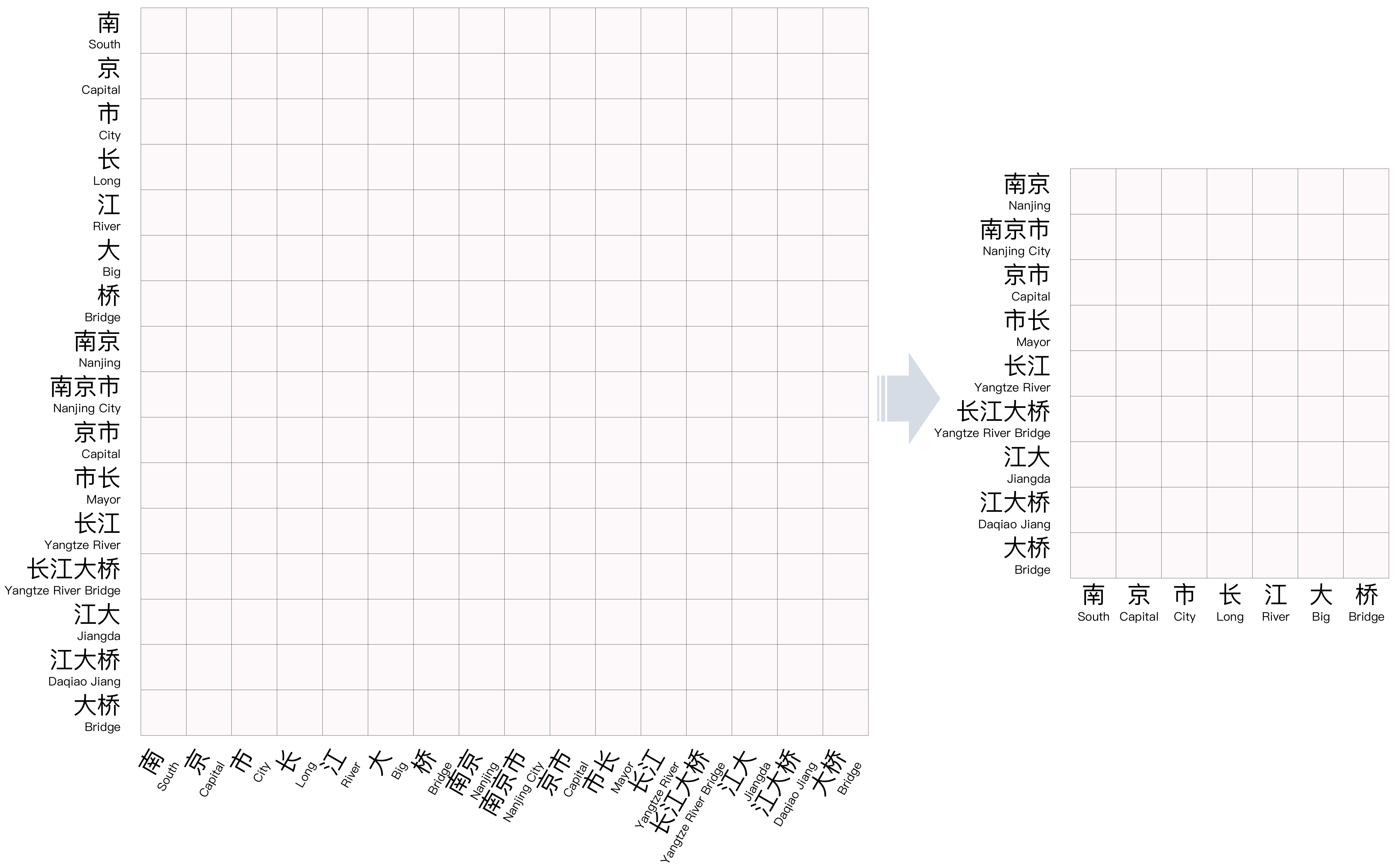}
    \label{fig:att-matrix}}
    
    \caption{FLAT and its attention matrix. (a) The input of FLAT is a flat lattice constructed by two positional encodings. The red one is the character encoding, and the green one is the word encoding. (b) The self-attention matrix of FLAT becomes huge when the sequence is long. The proposed InterFormer method simplifies the attention matrix significantly.}
    \label{fig:flat-attention}
    \vspace{-1em}
\end{figure}

As shown in Fig.~\ref{fig:flat}, FLAT models characters and words by introducing two positional encodings and constructing a set of flat lattices as the input of the model.
It solves the problems of blurred word boundaries and the lack of word semantics.
However, this method may match more words when dealing with long texts, resulting in long input sequences and more computational cost.
So FLAT struggles when dealing with sentences with lengths greater than 200.
More importantly, we argue that the computation between ``word-character'' and ``word-word'' in self-attention is not necessary (Fig.~\ref{fig:att-matrix}).
The reason is that the word representations containing global information are discarded at the decoding stage (Fig.~\ref{fig:flat}).

Another Transformer-based Chinese NER method is the character-based TENER~\cite{yan-etal-2019-tener}.
When using Transformer for Chinese NER, TENER proposed two optimizations: 1) The attention score is calculated using relative position encoding with orientation and distance awareness.
2) The attention score results can be calculated smoothly without scaling factors.
We implement \method using InterFormer with the help of TENER.
\method decouples lexicon fusion and context feature encoding, which has more advantages than FLAT in both accuracy and efficiency.

\section{Model Architecture}

\begin{figure*}
  \centering
  \includegraphics[scale=0.3]{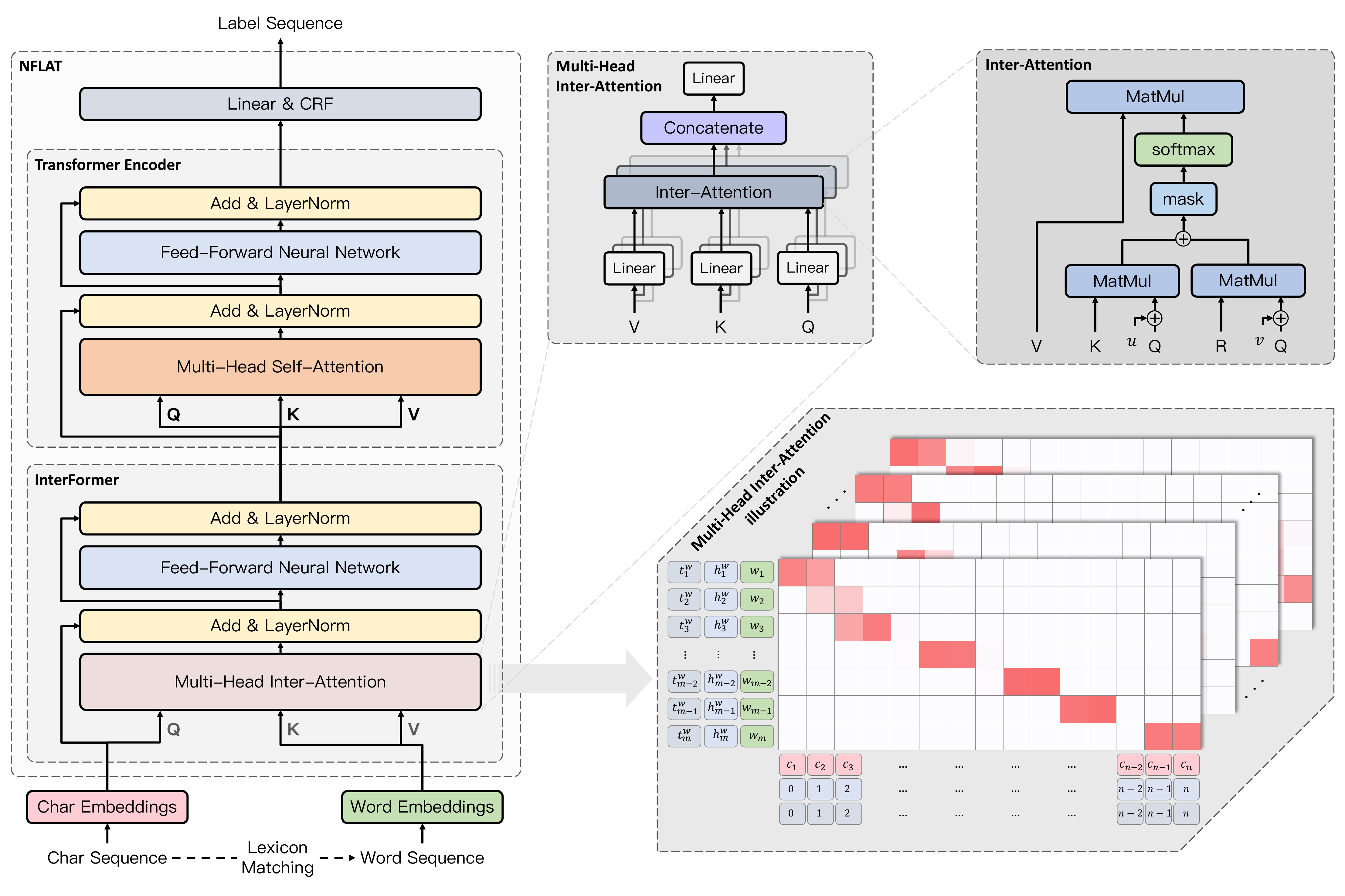}
  \caption{\label{fig:architecture}The overall architecture of \method.}
  \vspace{-1em}
\end{figure*}

For Chinese NER, \method has three main stages. 
The first stage applies InterFormer to fuse the words’ boundary and semantic information.
Then the second stage uses Transformer to encode the context with lexicon information.
Last, we use a Conditional Random Field (CRF) \cite{john-etal-2001-conditional} as the decoder to predict sequence labels.
The overall architecture of \method is shown in Fig.~\ref{fig:architecture}.

\subsection{InterFormer}
The proposed InterFormer method contains a multi-head inter-attention and a feed-forward neural network, as demonstrated in Fig.~\ref{fig:architecture}.
InterFormer aims to construct a non-flat-lattice and jointly model two sequences of characters and words with different lengths.
It enables the sequence of characters to fuse word boundaries and semantic information.

\subsubsection{Inter-Attention Module}
The Chinese character sequence,
$\mathcal{C} = \{c_1, c_2, ..., c_n\}$, and
word sequence,
$\mathcal{W} = \{w_1, w_2, ..., w_m\}$,
can be obtained by lexicon matching.
The input $\bm{Q}$, $\bm{K}$, $\bm{V}$ are obtained by the linear transformation of chars and words feature embedding:
\begin{equation}
    \left[\bm{Q}, \bm{K}, \bm{V}\right] = \left[\bm{X}^\mathcal{C}\bm{W}_q, \bm{X}^\mathcal{W}\bm{W}_k, \bm{X}^\mathcal{W}\bm{W}_v\right],
\end{equation}
where the token embeddings of character and word sequences, 
$\bm{X}^\mathcal{C} = \{\bm{x}_{c_1}, \bm{x}_{c_2}, ..., \bm{x}_{c_n}\}$
and
$\bm{X}^\mathcal{W} = \{\bm{x}_{w_1}, \bm{x}_{w_2}, ..., \bm{x}_{w_m}\}$,
are obtained by a word embedding lookup table. And each $\bm{W}$ is a learnable parameter.
In this paper, we use inter-attention to fuse lexicon information:
\begin{gather}
\operatorname{InterAtt}\left(\bm{A}, \bm{V}\right) = \operatorname{softmax}\left(\operatorname{mask}(\bm{A})\right)\bm{V},\\
\bm{A}_{ij} = \left(\bm{Q}_i + \bm{u}\right)^\top\bm{K}_j + \left(\bm{Q}_i + \bm{v}\right)^\top\bm{R}^*_{ij},\label{eq:att}
\end{gather}
where $1 \leq i \leq n$, $1 \leq j \leq m$.
The $\bm{u}$, $\bm{v}$ are learnable parameters.
Eq.~(\ref{eq:att}) is from Dai \textit{et al.}\cite{dai-etal-2019-transformer}.
\textit{mask()} is the inter-attention score mask of characters and words, which is 2-dimensional for a single batch and 3-dimensional for a multi-batch.
It is used to fill empty positions in the sequence with the value of $10^{-15}$, so that the attention weights of these positions are close to 0 when \textit{softmax()} normalization.
$\bm{R}^*_{ij}$ is calculated in a similar way to FLAT, using two  relative positions:
\begin{equation}
  \bm{R}_{ij} = \operatorname{ReLU}\left(\bm{W}_r\left(\bm{p}_{h^c_i - h^w_j} \oplus \bm{p}_{t^c_i - t^w_j}\right)\right),
\end{equation}
where $\bm{W}_r$ is a learnable parameter, $h$ and $t$ are the position numbers of the first and last characters of the word in the input text, and the superscripts $c$ and $w$ represent characters and words, respectively. $h^c_i - h^w_j$ represents the head position offset of the $i$-th character and the $j$-th word, $t^c_i - t^w_j$ represents the tail position offset of the $i$-th character and the $j$-th word, that is, the relative position.
The position encoding $\bm{p}$ generation method is proposed by Vaswani \textit{et al.}\cite{vaswani-etal-2017-attention}:
\begin{align}
  \bm{p}_{span}^{(2k)} &= \sin\left(\frac{span}{10000^{2k/d_{model}}}\right),\\
  \bm{p}_{span}^{(2k+1)} &= \cos\left(\frac{span}{10000^{2k/d_{model}}}\right),
\end{align}
where $span$ represents $h^c_i - h^w_j$ and $t^c_i - t^w_j$, $k$ is the $k$-th dimension, and $d_{model}$ is the hidden size.

\subsubsection{Multi-Head Inter-Attention}
In our preliminary experiments, we found that multi-head inter-attention can more effectively fuse lexicon information, and the information of different heads has complementary effects.
The multi-head inter-attention is calculated as follows:
\begin{gather}
	\text{Head}^{(s)} = \text{InterAtt}\left(\bm{X}^{\mathcal{C},(s)}, \bm{X}^{\mathcal{W},(s)}\right),\\
	\text{MultiHead}\left(\bm{X}^\mathcal{C}, \bm{X}^\mathcal{W}\right) = \left[\text{Head}^{(1)}, ..., \text{Head}^{(l)}\right],
\end{gather}
where $l$ is the number of inter-attention heads, $\text{Head}^{(s)}$ is the output result of the $s$-th inter-attention head on the character and word vector subspaces. $\bm{X}^{\mathcal{C},(s)}$ and $\bm{X}^{\mathcal{W},(s)}$ are the vector representations of characters and words in their subspaces.

\subsubsection{Feed-Forward Neural Network Module}
This paper refers to the design of the Transformer encoder. It implements the feed-forward neural network sub-module of the inter-attention encoder, using two fully connected layers:
\begin{equation}
  \text{FFN}\left(x\right) = \text{max}\left(0, \bm{x} \bm{W}_1 + \bm{b}_1\right)\bm{W}_2 + \bm{b}_2.
\end{equation}

In addition, we also use residual connection and layer normalization~\cite{ba-etal-2016-layer} in the above two sub-modules to speed up the convergence of network training and prevent the gradient vanishing problem:
\begin{gather}
    \text{output} = \text{LayerNorm}(\bm{X} + \text{SubModule}(\bm{X}))
\end{gather}
where $\bm{X}$ are the outputs of the above two sub-modules.

\subsection{Transformer Encoder}
After InterFormer, the character features are fused with the lexicon information.
Then, we use the Transformer encoder to encode the contextual information.
This part is inspired by Yan \textit{et al.}\cite{yan-etal-2019-tener}, in which the unscaled self-attention is found more suitable for NER.
In addition, the relative position encoding with orientation and distance awareness is adopted.
After this, we use the linear layer to project output into the label space and use CRF for decoding.

\section{Experiments}
We evaluate the proposed \method method using the F1 score (F1), precision (P), and recall (R) metrics, with a comparison of several character-word hybrid models.
At the same time, this section visually analyzes the inter-attention weights to verify the effectiveness of the proposed non-flat-lattice method.
Last, we also examine the complexity of \method and the flexibility of InterFormer and compare the effects of other lexicons and the use of pre-trained models.

\subsection{Experimental Settings}
\label{sec:exp-setting}
\paragraph{Data}
\input{tables/statistics_of_datasets}

This section evaluates the performance of \method on four mainstream Chinese NER benchmarking datasets, including three publicly available datasets, \textit{i.e.}, Weibo~\cite{peng-dredze-2015-named, he-sun-2017-f}, Resume~\cite{zhang-yang-2018-chinese} and MSRA~\cite{levow-2006-third}, and one licensed dataset, \textit{i.e.}, Ontonotes 4.0~\cite{weischedel-etal-2011-ontonotes}. Table~\ref{tab:statistics} shows the statistical information of these datasets.
Our models are all trained with a RTX 2080 Ti card.

\paragraph{Word Embedding}
The lighter structure of \method allows us to evaluate its performance on lexicons of different sizes.
We conducted experiments on three lexicons, YJ\footnote{\url{https://github.com/jiesutd/RichWordSegmentor}}~\cite{yang-etal-2017-neural-word}, LS\footnote{\url{https://github.com/Embedding/Chinese-Word-Vectors}}~\cite{li-etal-2018-analogical}, and TX\footnote{\url{https://ai.tencent.com/ailab/nlp/en/embedding.html}(v0.1.0)}~\cite{song-etal-2018-directional}.
Their statistics can be viewed in the appendix.

\paragraph{Hyper-parameter Settings}
This paper uses only one layer of InterFormer encoder and one layer of Transformer encoder to handle Chinese NER.
More details can be found in the appendix.

\input{tables/experiment_result}

\subsection{Experimental Results}
To compare the experimental results more reasonably, we use YJ, the most widely used in lexical enhancement methods, as the external lexicon.
We compare our \method method with almost all the lexical enhancement models, including Lattice-LSTM~\cite{zhang-yang-2018-chinese}, WC-LSTM~\cite{liu-etal-2019-encoding}, LR-CNN~\cite{gui-etal-2019-cnn}, LGN~\cite{gui-etal-2019-lexicon}, PLTE~\cite{mengge-etal-2020-porous}, SoftLexicon~\cite{ma-etal-2020-simplify}, and FLAT~\cite{li-etal-2020-flat}. 
The experimental results are shown in Table~\ref{tab:experiment-reslut}.
Note that we use TENER~\cite{yan-etal-2019-tener} and FLAT as the baseline models.
We can see that \method significantly improves the performance of TENER.
The overall F1 score on Weibo is increased by 3.77\%.
The F1 score obtained on Resume is increased by 0.58\%.
The F1 score on Ontonotes 4.0 is increased by 4.78\%, and the recall is 79.37\%.
For the MSRA dataset, the propose method increases the F1 score by 1.81\%, and achieves the best performance in terms of precision (P) and recall (R).
\method outperforms all the other methods, including FLAT, on all the datasets.
Without using other data augmentation methods and pre-trained language models, \method achieves state-of-the-art performance on the Weibo, Ontonotes 4.0, and MSRA datasets.

\subsection{Analysis of Inter-Attention}
\paragraph{Features}
The proposed inter-attention module can be understood as an interactive attention mechanism.
Self-attention expects all tokens in a sequence to be connected pairwise.
Inter-attention is different from self-attention, which expects to establish the connections between tokens belonging to two sequences of indeterminate length.
Similar attention mechanisms tend to appear in reading comprehension tasks~\cite{cui-etal-2017-attention}.
Unlike the existing attention mechanism, we draw on FLAT to design the relative position encoding to enable it to discover the potential relationship between the tokens in two sequences.
The original intention of this paper is to establish a connection between character sequences and word sequences so that the character sequence can fuse the boundary and semantic information of words.

\begin{figure*}
  \centering
  \includegraphics[scale=0.55]{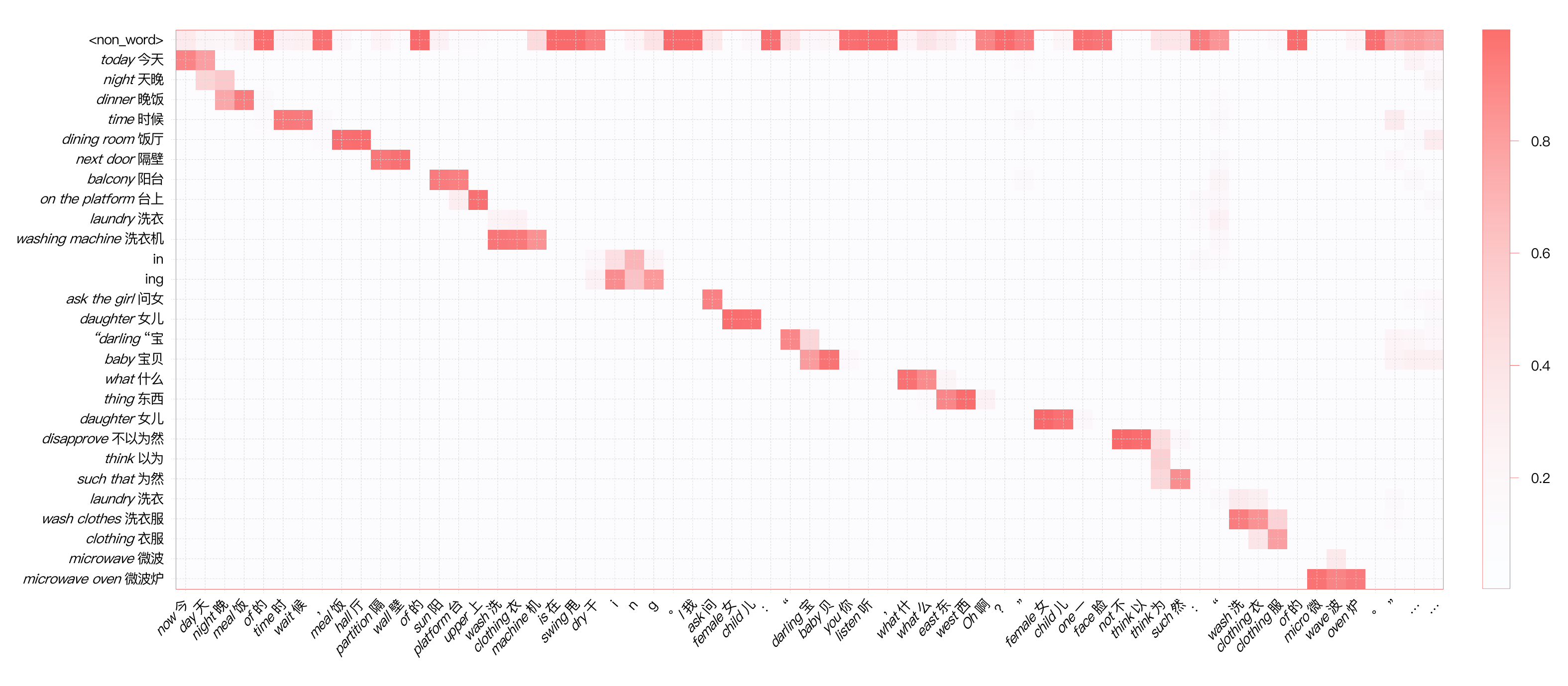}
  \caption{\label{fig:att-visual}Visualization of the proposed inter-attention module.}
\end{figure*}

\paragraph{Auto-discovery}
As shown in Fig.~\ref{fig:att-visual}, we visualize the attention weights obtained by the inter-attention module.
The heatmap will be a massive square if the FLAT method is used.
It can be seen that the inter-attention module can effectively learn the correlation between characters and words.
It is worth noting that the word ``洗衣机 \textit{washing machine}'' in this sentence is given a higher weight than those of the characters ``洗 \textit{washing}'' and ``衣 \textit{clothing}'' than ``洗衣 \textit{laundry}''.
The same is true for ``今天 \textit{today}'', ``天晚 \textit{night}'' and ``晚饭 \textit{dinner}''; ``微波炉 \textit{microwave oven}'' and ``微波 \textit{microwave}''; ``ing'' and ``in''; ``不以为然 \textit{disapprove}'', ``以为 \textit{think}'' and ``为然 \textit{such that}''; ``洗衣服 \textit{wash clothes}'' and ``衣服 \textit{clothing}''.
Therefore, inter-attention can automatically discover valid words and increase their weights while reducing the weights of invalid words.

\paragraph{Auto-suppression}
\label{sec:extra-tag}
It can also be seen from Fig.~\ref{fig:att-visual} that the proposed method adds the \textit{\textless non\_word\textgreater} tag.
This tag establishes the connections between punctuation marks, single characters, or characters that do not match the lexicon.
According to the visualization result, the attention weights of punctuation marks such as ``，'', ``。'', ``：'', ``…'', ``？'' are concentrated on the \textit{\textless non\_word\textgreater} tag.
Besides, the stop words such as ``的 \textit{of}'', ``在 \textit{is}'', ``你 \textit{you}'' are more concerned with the \textit{\textless non\_word\textgreater} tag.
There are also characters not matched to the lexicon, which are also assigned to the weights of the \textit{\textless non\_word\textgreater} tag, such as ``甩 \textit{swing}'', ``干 \textit{dry}'', ``一 \textit{one}'', ``脸 \textit{face}''.
So inter-attention can automatically suppress irrelevant characters and reduce noise information.

\paragraph{Interpretability}
We further examine the focus of other attention heads for multi-head interactive attention.
Some of them pay more attention to the first character, and some have closer attention to the last character.
This provides favorable evidence for interpreting the proposed method.
The inter-attention module can be confirmed to learn the boundary relationship between Chinese characters and words very well.
Moreover, it incorporates lexicon information to improve the performance of Chinese NER.

This paper does not purposely design ablation experiments for the inter-attention module.
Since the superstructure of \method is TENER, it can be directly compared with the experimental results of the baseline TENER method.
From Table~\ref{tab:experiment-reslut}, we can see that \method outperforms TENER, which is since the input of the self-attention encoder has fused lexicon information through the inter-attention encoder.
This verifies the effectiveness of the non-flat-lattice architecture constructed by the inter-attention mechanism for performance boosting.

\subsection{Complexity Analysis}
\label{sec:compute}
\begin{figure}[!t]
    \centering
    \subfloat[]{\includegraphics[width=3in]{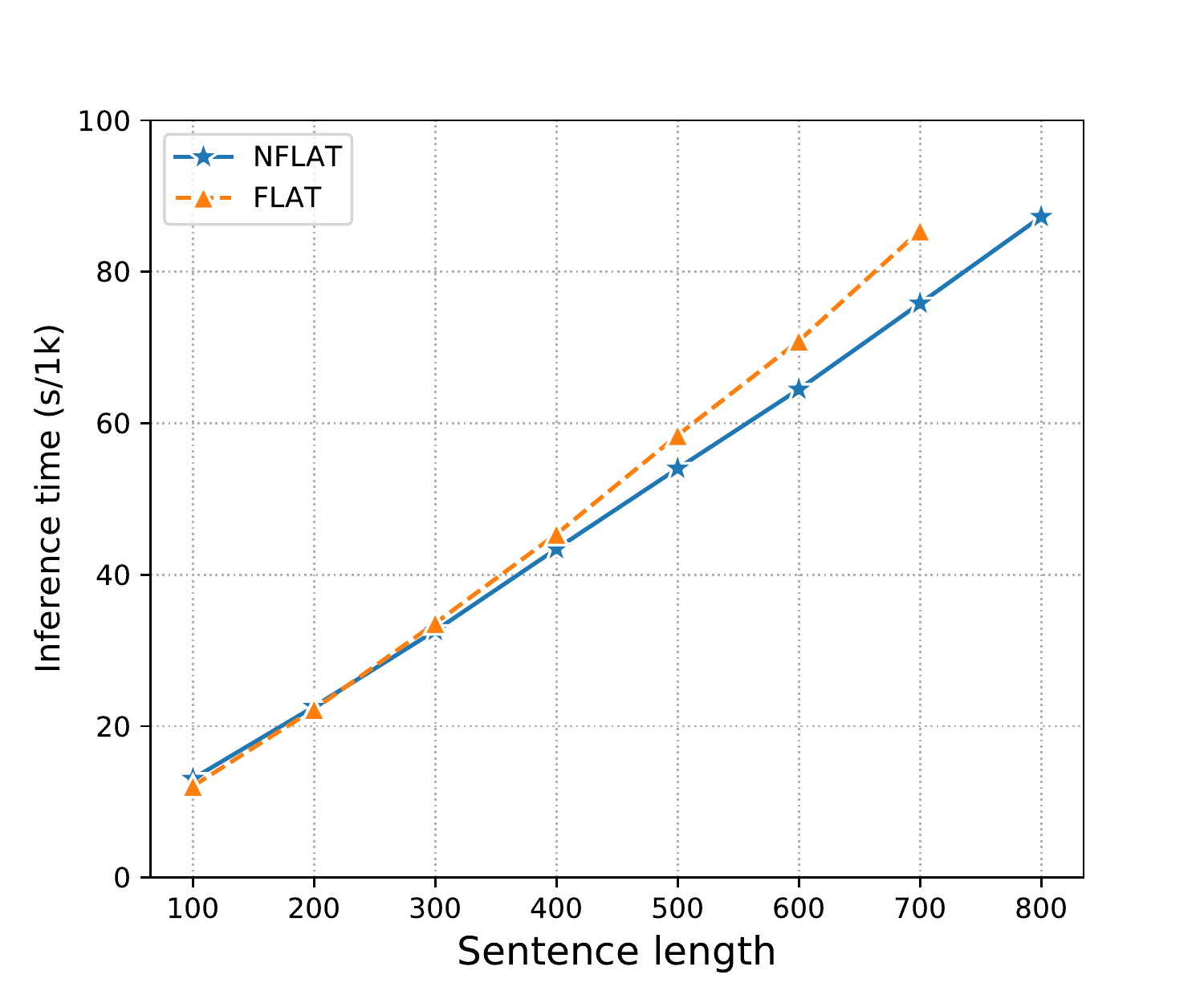}
    \label{fig:efficiency}}
    
    \subfloat[]{\includegraphics[width=3in]{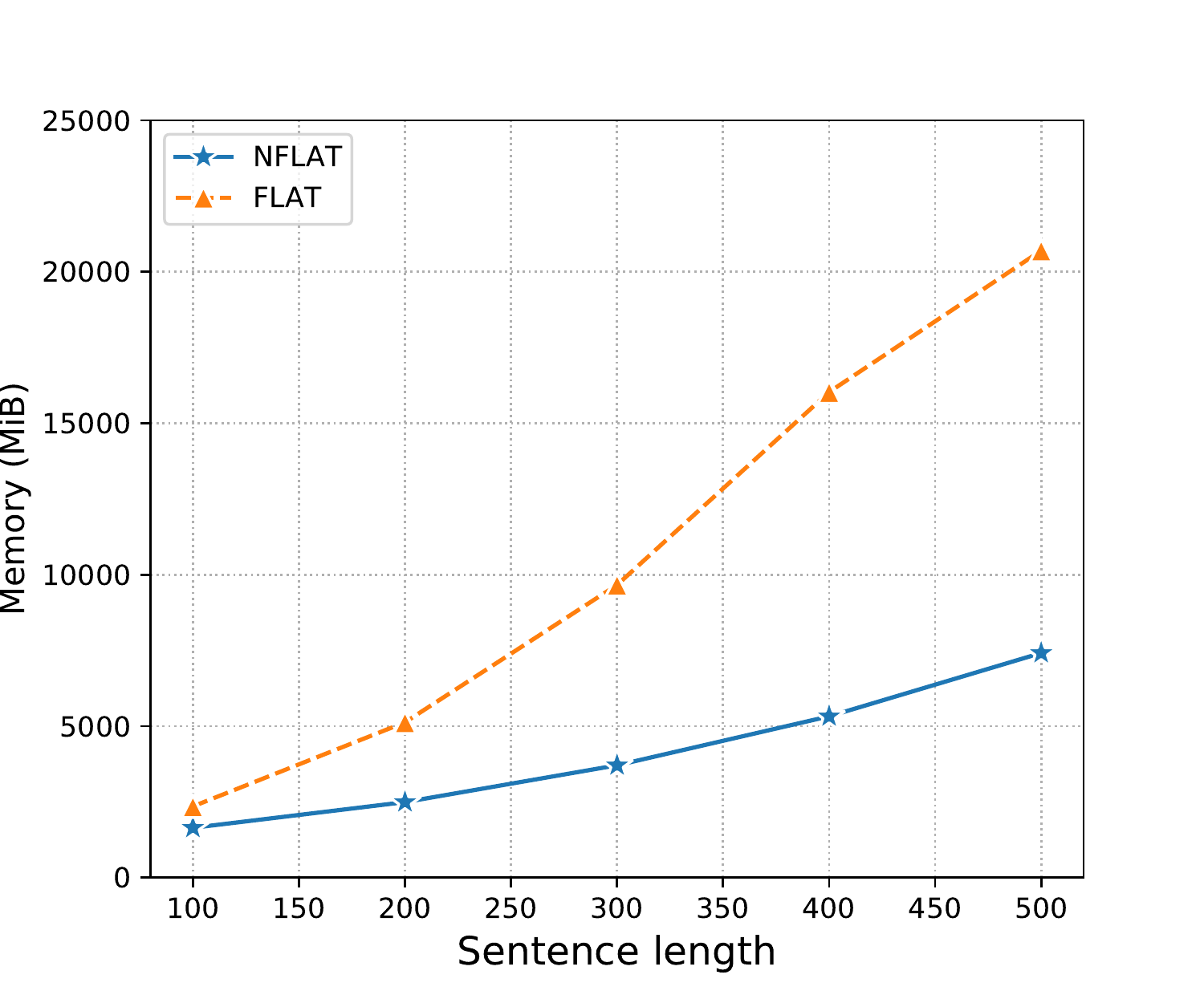}
    \label{fig:memory}}
    
    \caption{A comparison between FLAT and \method in terms of efficiency and memory usage, evaluated at different sentence lengths. (a) Efficiency. (b) Memory usage.}
    \label{fig:complexity}
\end{figure}

\paragraph{Time Complexity}
Let us only consider the attention module. \method includes the inter-attention of characters and words($\mathcal{O}(nmd)$), and the self-attention($\mathcal{O}(n^2d)$) of context, so the total complexity is ($\mathcal{O}((n+m)nd)$). The lattice attention complexity of FLAT is ($\mathcal{O}((n+m)^2d)$).
As the length $n$ of the Chinese character sequence increases, the size $m$ of the matched word sequence must also increase, making the computational complexity difficult to estimate.
Fig.~\ref{fig:efficiency} compares \method and FLAT in the inference speed on one RTX 3090 card.
We choose 1,000 sentences for each length, set the batch size to 1 and calculate the overall time (s/1k) used for processing all the 1k sentences. 
We can see that FLAT and \method perform similarly when the sentence length is lower than 400. After that, \method becomes faster than FLAT as the sentence length increases.
When the sentence length exceeds 700, FLAT can not run on the GPU card due to its high memory usage.

\paragraph{Space Complexity}
We also compare FLAT and \method in terms of memory usage when processing sentences with different lengths in Fig.~\ref{fig:memory}.
In FLAT, the self-attention computation requires multiplying two $(n+m) \times d$ matrices, so the space complexity is $\mathcal{O}\left((n+m)^2\right)$.
The space complexity of self-attention and inter-attention in \method is $\mathcal{O}\left(n^2\right)$ and $\mathcal{O}\left(nm\right)$, respectively.
The larger $n$ is, the larger $m$ is. So the memory usage of \method is almost half of that of FLAT.
As shown in the figure, FLAT occupies more and more memory as the sentence length increases until it stops working on the RTX 3090 card.
On the other hand, \method has a more stable memory occupancy rate.
It still works after the sentence length exceeds 1000 and can improve the utilization of the device during the training stage.

\paragraph{Model Size}
With different hyper-parameter settings, the parameter size of \method is between 0.43M and 1.27M.

\input{tables/experiment_result_other_lex}
\input{tables/ablation_experiment}

\subsection{Flexibility of InterFormer}
InterFormer can be extracted from \method as a separate module.
Compared with FLAT, it is more flexible because it does not rely on Transformer.
It can be connected to LSTM, Bi-LSTM, CNN, Transformer, and even pre-trained models according to different tasks at the fusion layer.
InterFormer can also refer to the design of LEBERT~\cite{liu-etal-2021-lexicon} and add it to the middle layer of a pre-trained model.
In addition, because it can establish a relationship between two sequences of different lengths, we can directly use a layer of InterFormer to model ``characters'' and ``characters + words''.
Although this method is not as efficient as \method, it still has a decent performance.

\subsection{How Inter-Attention Performs}
The key to the success and effectiveness of NFLAT lies in the inter-attention module of InterFormer. Note that inter-attention could be easily misunderstood as cross-attention because they are computed similarly. However, inter-attention additionally uses the relative position encoding information to associate character-word pairs.

To verify the effectiveness of inter-attention, we perform two ablation experiments. First, we remove the relative position encoding, which makes inter-attention more similar to cross-attention. We denote this method as `-RPE'. Second, we remove the \textit{\textless non\_word\textgreater} tag to evaluate the performance of using this extra tag, which is denoted as `-TAG'. The results of the ablation experiments are reported in Table~\ref{tab:ablation_experiment}.

Based on the ablation experimental results, we can see that the performance of NFLAT degrades when removing the relative position encoding. The `-RPE' method, which is more similar to cross-attention, loses word boundary information. It simply fuses the semantic information of words, so the performance is only marginally improved as compared with TENER. But it still performs significantly worse than FLAT. We also test the performance of NFLAT by retaining the relative position encoding but removing the auxiliary tags, denoted as `-TAG'. Again, the performance of the `-TAG' method is degraded. This is because inter-attention introduces additional noisy information when some characters may not have matching words. The self-attention module of FLAT forces these characters to pay more attention to themselves. NFLAT resolves this issue by using auxiliary tags thus inter-attention pays more attention to these tags.

\subsection{Compatibility with Other Lexicons and Pre-trained Models}
\label{sec:other-lex}
Since \method uses the lighter InterFormer as a module for fusing lexicon information, it is possible to use a larger and richer lexicon.
We additionally evaluate the performance of \method on two larger lexicons, LS and TX, and compare other models using them.
The experimental results are reported in Table~\ref{tab:experiment-reslut-other-lex}.
As the size of the lexicon increases, the performance of \method improves.
So a larger and richer lexicon is beneficial for \method.
\method outperforms FLAT and CGN~\cite{sui-etal-2019-leverage} when we use LS as the lexicon.
\method performs better than Lattice-LSTM when they use the TX lexicon.
\method has the best performance when a larger scale lexicon, TX, is used.

Furthermore, \method can easily integrate the pre-trained model into the embedding layer.
In this paper, we use \textit{BERT-wwm} released by~\cite{cui-etal-2021-pretrain}.
It can be seen from Table~\ref{tab:experiment-reslut-other-lex} that \method further improves the performance of the pre-trained model `BERT+CRF'.

\section{Conclusion}
We presented a novel non-flat-lattice InterFormer module, as well as the \method architecture, for Chinese NER with lexical enhancement.
\method decouples FLAT and divides Chinese NER into two stages: lexicon fusion and context encoding, which have apparent performance advantages and efficiency. 
InterFormer can flexibly associate two indefinite-length sequences and achieve significant performance boosting for Chinese NER.
InterFormer breaks the stereotype of the self-attention mechanism.
We believe that it has certain universality and can be extended to any task that requires joint modeling of two information sequences.
It is instrumental in multi-modal tasks, such as text and image sequences or speech and text sequences.


\bibliographystyle{IEEEtran}
\bibliography{anthology,custom}

\newpage

\begin{IEEEbiography}[{\includegraphics[width=1in,height=1.25in,clip,keepaspectratio]{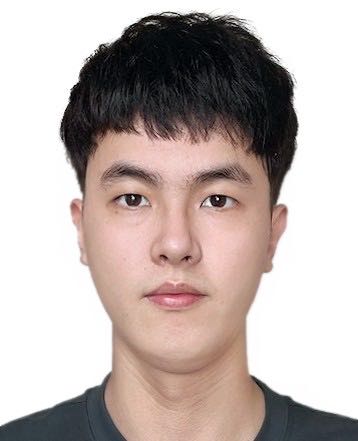}}]{Shuang Wu} is a master student at the School of
Artificial Intelligence and Computer Science, Jiangnan University, Wuxi, China. His research interests include natural language processing, data mining and machine learning.

\end{IEEEbiography}

\begin{IEEEbiography}[{\includegraphics[width=1in,height=1.25in,clip,keepaspectratio]{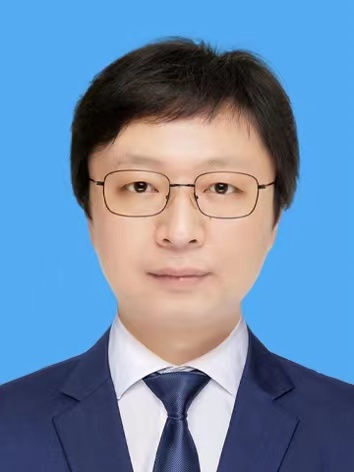}}]{Xiaoning Song}
received the BSc degree in computer
science from Southeast University, Nanjing, China,
in 1997, the MSc degree in computer science from
the Jiangsu University of Science and Technology,
Zhenjiang, China, in 2005, and the PhD degree
in pattern recognition and intelligence system from
the Nanjing University of Science and Technology,
Nanjing, in 2010. He was a visiting researcher
with the Centre for Vision, Speech, and Signal
Processing, University of Surrey, Guildford, UK,
from 2014 to 2015. He is currently a full Professor
with the School of Artificial Intelligence and Computer Science, Jiangnan University, Wuxi,
China. His research interests include pattern recognition, machine
learning, and computer vision.
\end{IEEEbiography}

\begin{IEEEbiography}[{\includegraphics[width=1in,height=1.25in,clip,keepaspectratio]{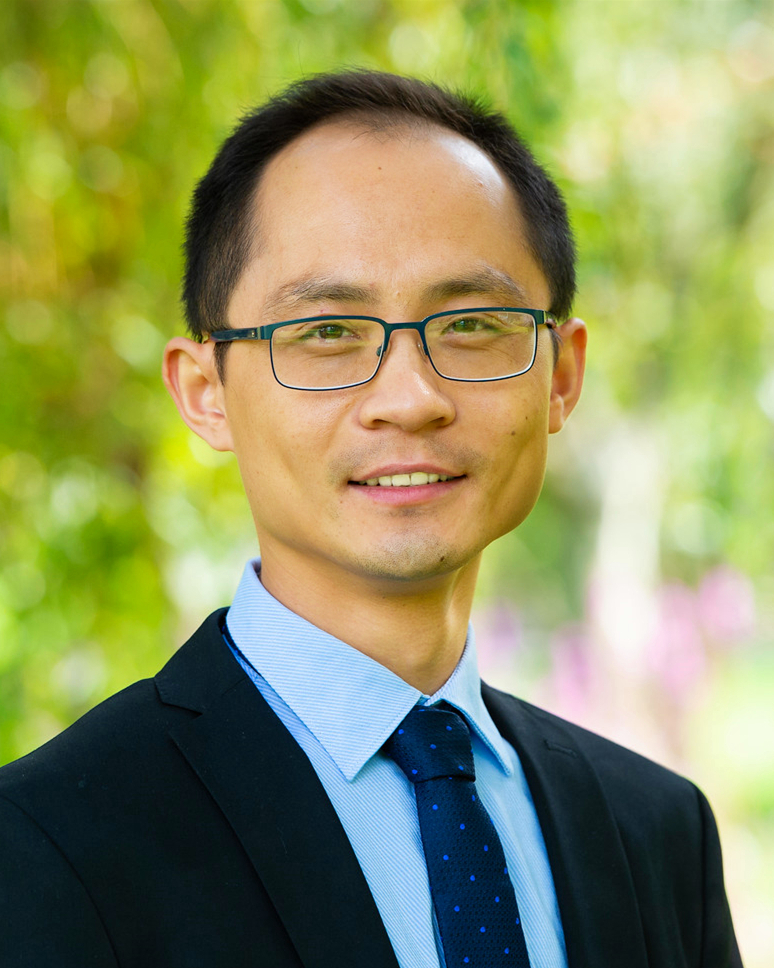}}]{Zhen-hua Feng} (S'13-M'16) received the Ph.D. degree from the Centre for Vision, Speech and Signal Processing (CVSSP), University of Surrey, Guildford, U.K. in 2016. He is currently a Lecturer in Computer Vision and Machine Learning at the Department of Computer Science, University of Surrey. Before this, he was a senior research fellow at CVSSP. His research interests include pattern recognition, machine learning and computer vision. He has published more than 60 research papers in top-tier conferences and journals such as TPAMI, IJCV, CVPR, ICCV, IJCAI, ACL, TCYB, TIP, TIFS, TCSVT, TBIOM, ACM TOMM, Information Sciences, Pattern Recognition, etc. 

He served as Area Chair for BMVC 2021 and Senior Programme Committee member for IJCAI 2021. He currently serves on the Editorial Board of Complex and Intelligent Systems. He received the 2017 European Biometrics Industry Award from the European Association for Biometrics (EAB) and the Best Paper Award for Commercial Applications from the AMDO2018 conference.
\end{IEEEbiography}

\begin{IEEEbiography}[{\includegraphics[trim=15mm 30mm 15mm 5mm, clip, width=1in, height=1.25in]{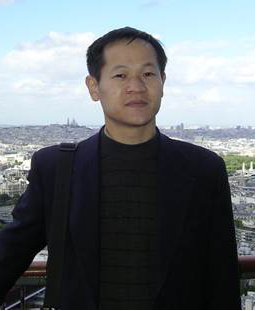}}]{Xiao-jun Wu}
received the BSc degree in mathematics from Nanjing Normal University, Nanjing,
China, in 1991, and the M.S. and Ph.D. degrees in
pattern recognition and intelligent systems from the
Nanjing University of Science and Technology, Nanjing, in 1996 and 2002, respectively. He is currently
a Distinguished Professor of Artificial Intelligence
and Pattern Recognition with Jiangnan University,
Wuxi, China. His current research interests include
pattern recognition, computer vision, fuzzy systems,
neural networks, and intelligent systems.
\end{IEEEbiography}

\appendix

\section{Appendix}

\subsection{Statistics of Lexicons}
Table~\ref{tab:lex_statistics} shows the statistics of the external lexicons used in this paper, mainly including their total words, single-character words, two-character words, three-character words, and words with more characters. 
Also, we list the dimensions of the word vectors in each lexicon.

We list the maximum and average character lengths for each dataset sample in Table~\ref{tab:word-match}.
It also shows the lengths of words matched on each lexicon for the four datasets.
It can be found that as the size of the lexicon increases, the length of matching words also becomes longer.
So it makes FLAT challenging to deal with large-scale lexicons, and \method can make up for this shortcoming and bring better performance.

\input{tables/statistics_of_lexicons}
\input{tables/word_match}

\subsection{Hyper-parameters Selection}
\label{app:hyper-parameters-select}
We manually selected parameters on the two large-scale datasets, including Ontonotes 4.0 and MSRA. We used the SMAC~\cite{hutter-etal-2011-sequential} algorithm to search for the best hyper-parameters for the two small datasets, Weibo and Resume. 
The range of parameters is listed in Table~\ref{tab:range_of_hyper-parameters}.

\subsection{Case Study}
Tables~\ref{tab:case_1} - \ref{tab:case_2} show several case studies comparing the results of FLAT and \method on the Ontonotes 4.0 test data.

The results in Table~\ref{tab:experiment-reslut} show that \method significantly improves recall due to its incorporation of lexicon information.
Nevertheless, it also brought a slight decrease in precision.
It can be seen from Table~\ref{tab:case_1} that there are some problems with missing gold labels in the Ontonotes 4.0 test data.
In sentence 1, ``九龙街 \textit{Kowloon Street}'' is not annotated, but both FLAT and \method can correctly identify it.
In sentence 2, ``梁 \textit{Liang}'' is missing the gold label, and both FLAT and \method recognize this entity, while \method’s results are more accurate than FLAT.
In sentence 3, neither the gold label nor FLAT annotates ``萨国 \textit{Republic of El Salvador}'', only the \method prediction is correct.
Accordingly, if the test data is correctly annotated, we believe that \method will get a higher F1 with its excellent recall.

The advantages of \method over FLAT can be analyzed from the case studies in Table~\ref{tab:case_2}.
In sentence 1, FLAT truncates the 13-character long entity into a GPE entity and an ORG entity, while \method does not.
It shows that FLAT is not as good as \method in identifying the boundaries of long entities.
In sentences 2 and 3, FLAT increases the boundary range of the entity.
Nevertheless, the \method gives the correct answer due to the effect of the ``\text{<non\_word>}'' tag in the \method can automatically disable noise words.
In sentence 4, ``津巴布韦 \textit{Zimbabwe}'' (GPE) and ``斯 · 埃万 \textit{Si Aiwan}'' (PER) were not predicted by FLAT, while \method effectively predicted them.
The above suggests that \method's ability to incorporate lexicon information is better than FLAT in some aspects.

\subsection{Learning Curves}
\label{app:learning-curves}
\begin{figure*}
    \centering
    \subfloat[]{\includegraphics[width=3in]{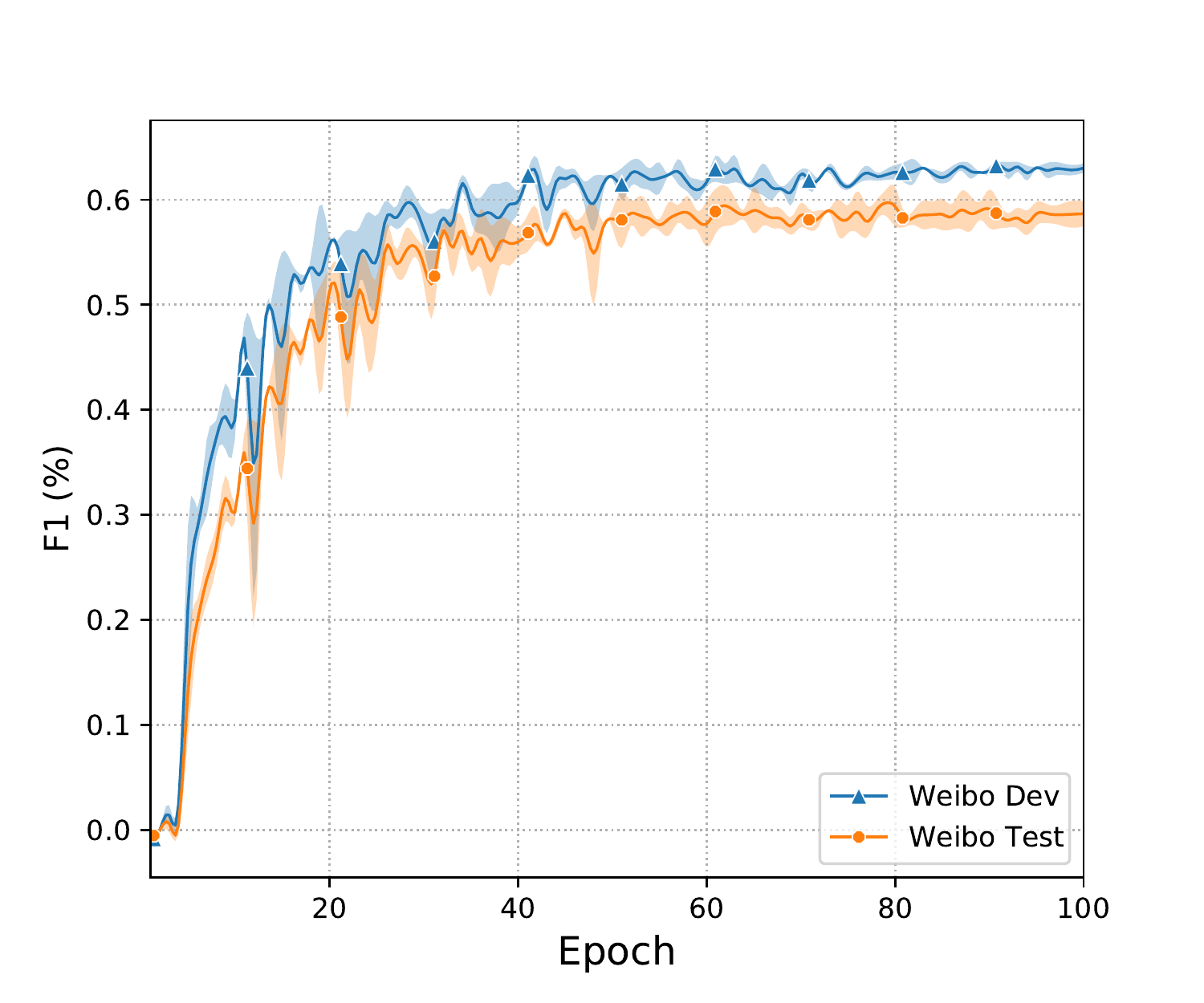}}
    \subfloat[]{\includegraphics[width=3in]{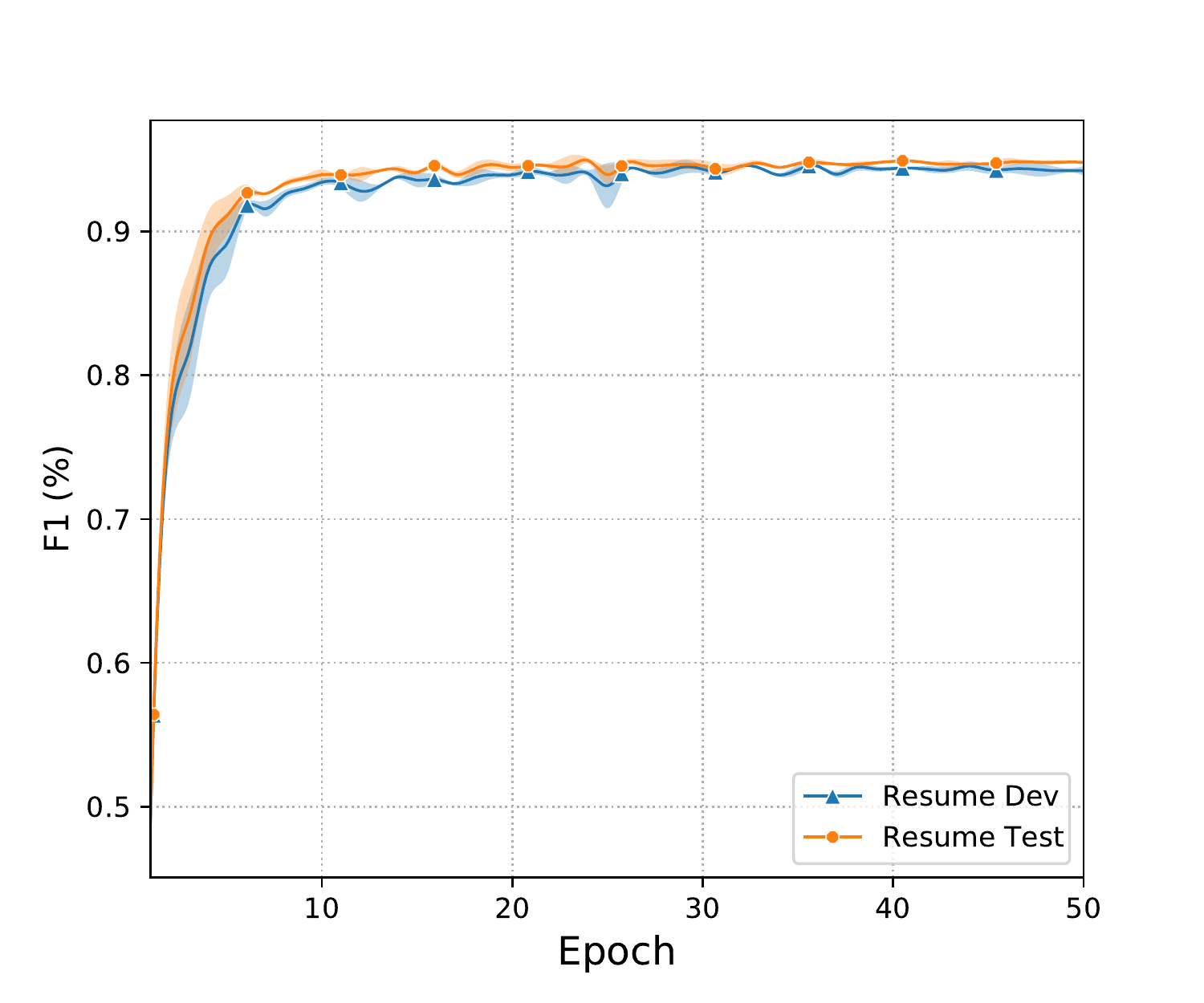}}
    
    \subfloat[]{\includegraphics[width=3in]{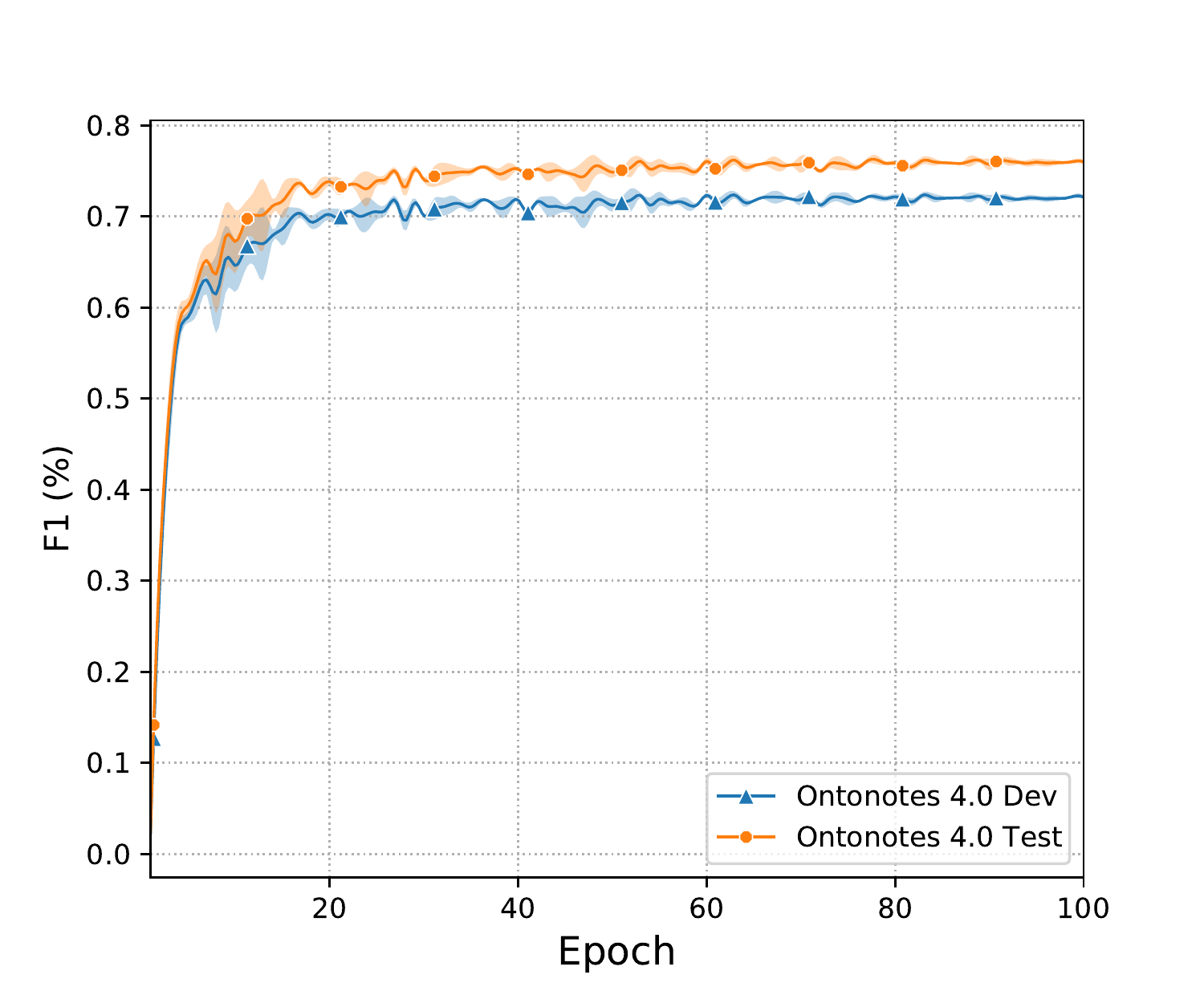}}
    \subfloat[]{\includegraphics[width=3in]{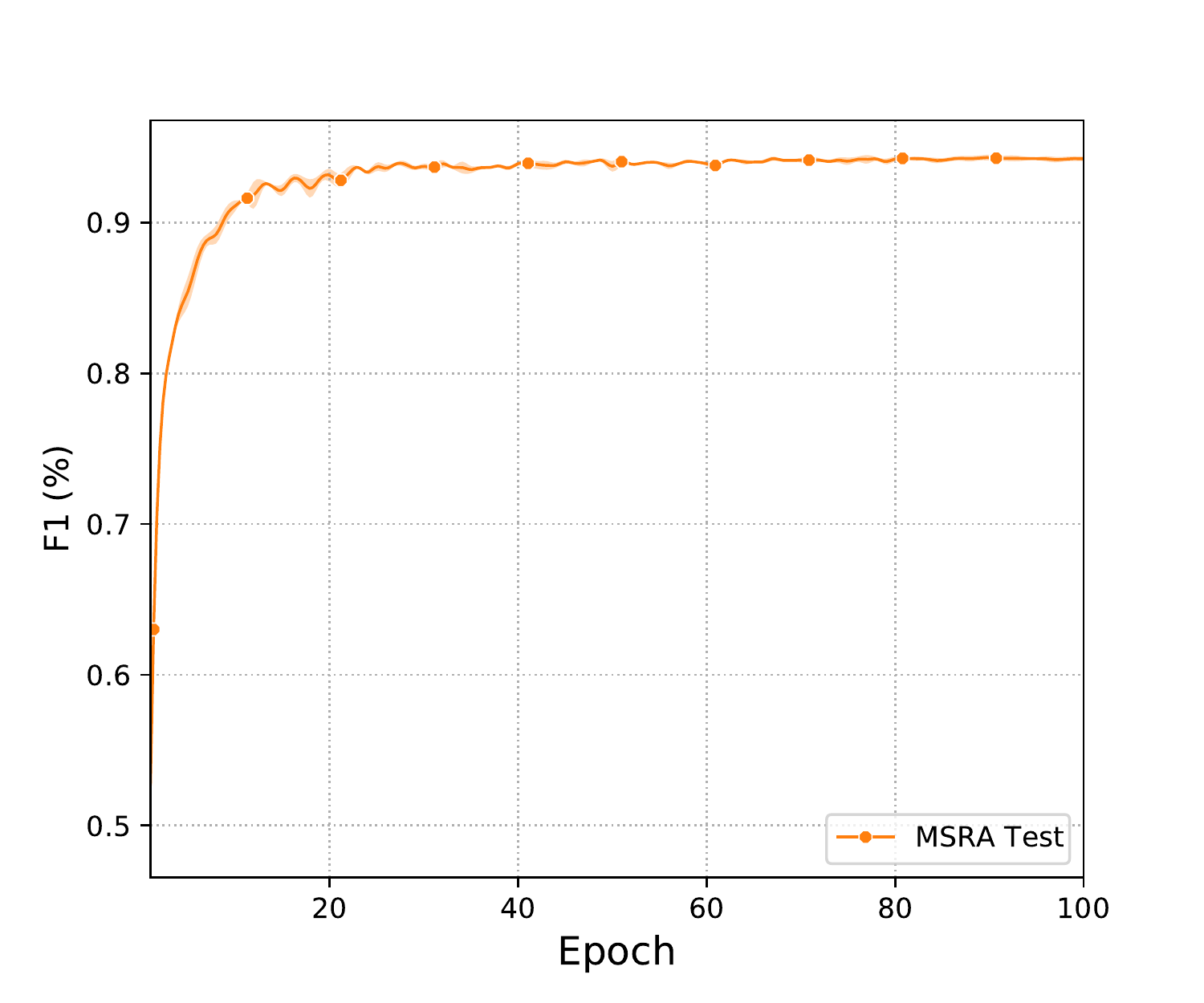}}
    
    
    \caption{Learning curves on four datasets. The lexicon used is YJ. There is no dev set in the MSRA dataset. And the result with averages and standard deviations across 3 seeds. (a) Learning curve on Weibo. (b) Learning curve on Resume. (c) Learning curve on Ontonotes 4.0. (d) Learning curve on MSRA.}
    \vspace{-2em}\label{fig:learning-curves}
\end{figure*}

We evaluate the performance of \method on each dataset with different seeds.
The results are shown in Fig.~\ref{fig:learning-curves}.
It can be seen that its results on Resume, Ontonotes 4.0, and MSRA become more stable with increasing training iterations.
However, the performance fluctuates significantly during the training process on Weibo.
Weibo data is relatively small from the field of social media, and the text is colloquial.
There are network words, noise information, and differences in the data, so the fluctuation is relatively large.
This phenomenon also appears in other models, such as FLAT.

\subsection{Societal Impacts}
\label{app:societal_impacts}
The weights of word embeddings and pre-trained models used in this article are all from assets released by others.
It is difficult to guarantee that this information is free from bias regarding gender, race, abuse, violence, physical ability, etc.
It is mainly related to the corpus used for training.
We do not avoid these biases, but they generally do not occur in our experimental datasets.

In addition to this, the datasets we use are collected from the web or news. Some data may contain sensitive content involving privacy, politics, and offensiveness. We have avoided presenting this kind of data in our case studies.

\subsection{Limitations}
\label{app:limitations}
The NER method used in this paper is limited to Chinese.
We did not evaluate \method performance in other languages.
Although we think this method might not work for English, it might work for Japanese, Thai, etc., similar to Chinese.
A feature of these languages is that there is no clear word boundary between words, just like the use of spaces to separate words in English sentences.
Besides, the performance of our method is limited by the quality, size, richness, and domain of the lexicon.
These factors will affect the performance of NER to varying degrees, as Section~\ref{sec:other-lex} describes the results obtained with different richness or sizes.

\input{tables/hyper_parameters_range}
\input{tables/study_case_1}
\input{tables/study_case_2}

\end{CJK}
\end{document}

%% file: tables/statistics_of_datasets.tex
\begin{table}[htp]
  \caption{Statistics of the benchmarking datasets.}
  \label{tab:statistics}
  \centering
  \begin{center}
  \begin{tabular}{lcccc}
    \toprule
	Datasets & Items & Train & Dev & Test\\
    \midrule
	\multirow{2.5}{*}{Weibo} & Sentences & 1.35k & 0.27k & 0.27k\\
	 & Entities  & 1.89k & 0.39k & 0.42k\\
	\cmidrule(r){2-5}
	\multirow{2.5}{*}{Resume} & Sentences & 3.8k  & 0.46k & 0.48k\\
	 & Entities  & 1.34k & 0.16k & 0.16k\\
	\cmidrule(r){2-5}
	\multirow{2.5}{*}{MSRA} & Sentences & 46.4k & - & 4.4k\\
	 & Entities  & 74.8k & - & 6.2k\\
	\cmidrule(r){2-5}
	\multirow{2.5}{*}{OntoNotes 4.0} & Sentences & 15.7k & 4.3k  & 4.3k\\
	 & Entities  & 13.4k & 6.95k & 7.7k\\
    \bottomrule
  \end{tabular}
  \end{center}
\end{table}

%% file: tables/experiment_result.tex
\begin{table*}
  \caption{A comparison with other lexical enhancement methods. YJ lexicon is used here.}
  \label{tab:experiment-reslut}
  \centering
  \begin{tabular}{lccc|ccc|ccc|ccc}
    \toprule
	\multirow{2.5}{0.1\textwidth}{Models} & \multicolumn{3}{c}{Weibo} & \multicolumn{3}{c}{Resume} & \multicolumn{3}{c}{Ontonotes 4.0} & \multicolumn{3}{c}{MSRA} \\
	\cmidrule(r){2-13}
	 & NE & NM & Overall & P & R & F1 & P & R & F1 & P & R & F1 \\
    \midrule
	Lattice-LSTM  & 53.04 & 62.25 & 58.79 & 94.81 & 94.11 & 94.46 & 76.35 & 71.56 & 73.88 & 93.57 & 92.79 & 93.18 \\
    WC-LSTM       & 53.19 & \textbf{67.41} & 59.84 & 95.27 & 95.15 & 95.21 & 76.09 & 72.85 & 74.43 & 94.58 & 92.91 & 93.74 \\
    LR-CNN        & 57.14 & 66.67 & 59.92 & 95.37 & 94.84 & 95.11 & 76.40 & 72.60 & 74.45 & 94.50 & 92.93 & 93.71 \\
    LGN           & 55.34 & 64.98 & 60.21 & 95.28 & 95.46 & 95.37 & 76.13 & 73.68 & 74.89 & 94.19 & 92.73 & 93.46 \\
	PLTE          & 53.55 & 64.90 & 59.76 & 95.34 & 95.46 & 95.40 & 76.78 & 72.54 & 74.60 & 94.25 & 92.30 & 93.26 \\
	SoftLexicon   & 59.08 & 62.22 & 61.42 & \textbf{95.71} & \textbf{95.77} & \textbf{95.74} & \textbf{77.13} & 75.22 & 76.16 & 94.73 & 93.40 & 94.06 \\
	\midrule
	TENER         &   -   &   -   & 58.17 &   -   &   -   & 95.00 &   -   &   -   & 72.43 &   -   &   -   & 92.74 \\
	FLAT          &   -   &   -   & 60.32 &   -   &   -   & 95.45 &   -   &   -   & 76.45 &   -   &   -   & 94.12 \\
    \rowcolor{lime!20}
	NFLAT         & \textbf{59.10} & 63.76 & \textbf{61.94} & 95.63 & 95.52 & \textbf{95.58} & 75.17 & \textbf{79.37} & \textbf{77.21}& \textbf{94.92} & \textbf{94.19} & \textbf{94.55}\\ 
    \bottomrule
  \end{tabular}
\end{table*}

%% file: tables/experiment_result_other_lex.tex
\begin{table*}
  \caption{Performance evaluation on other lexicons or pretrained models. The result marked with $\dagger$ is what we get by running the open source code. The results marked with $*$ are from Li \textit{et al.}\cite{li-etal-2020-flat}. And \textit{With Pre.} indicates whether to use a pre-trained model. (\%)}
  \label{tab:experiment-reslut-other-lex}
  \centering
  \begin{center}
  \begin{tabular}{lcccccc}
    \toprule
	Models & Lexicon & With Pre. & Weibo & Resume & Ontonotes & MSRA\\
    \midrule
	CGN   & LS & \XSolidBrush & 63.09 & \ \ 94.12$^*$ & 74.79 & 93.47 \\
    FLAT  & LS & \XSolidBrush & 63.42 & 94.93 & 75.70 & 94.35 \\
    \rowcolor{lime!20}
    NFLAT & LS & \XSolidBrush & \textbf{63.48} & \textbf{95.20} & \textbf{75.86} & \textbf{94.73} \\
	\midrule
	Lattice-LSTM   & TX(v0.1.0) & \XSolidBrush & \ \ 60.79$^\dagger$ & \ \ 94.98$^\dagger$ & \ \ 75.79$^\dagger$ & \ \ 94.20$^\dagger$ \\
    \rowcolor{lime!20}
    NFLAT & TX(v0.1.0) & \XSolidBrush & \textbf{66.40} & \textbf{95.62} & \textbf{76.80} & \textbf{94.76} \\
	\midrule
    BERT + CRF     & - & \CheckmarkBold & \ \ 68.20$^*$ & \ \ 95.53$^*$ & \ \ 80.14$^*$ & \ \ 94.95$^*$ \\
    \rowcolor{lime!20}
	BERT + NFLAT   & TX(v0.1.0) & \CheckmarkBold & \textbf{71.04} & \textbf{96.86} & \textbf{82.78} & \textbf{96.40} \\
    \bottomrule
  \end{tabular}
  \end{center}
\end{table*}

%% file: tables/ablation_experiment.tex
\begin{table}
  \caption{Ablation experimental results. We remove the relative position encoding in Eq.~(\ref{eq:att}) and the extra tags mentioned in Section~\ref{sec:extra-tag}, which are noted by `-RPE' and `-TAG', respectively. (\%)}
  \label{tab:ablation_experiment}
  \centering
  \begin{center}
  \begin{tabular}{lcccccc}
    \toprule
	Models & Weibo & Resume & Ontonotes & MSRA\\
    \midrule
    TENER & 58.17 & 95.00 & 72.43 & 92.74 \\
    FLAT & 60.32 & 95.45 & 76.45 & 94.12 \\
	\midrule
    \rowcolor{lime!20}
    NFLAT & \textbf{61.94} & \textbf{95.58} & \textbf{77.21} & \textbf{94.55} \\
	\quad - RPE   & 58.24 & 95.18 & 73.58 & 93.03 \\
    \quad - TAG   & 59.51 & 95.46 & 76.41 & 93.66 \\
    \bottomrule
  \end{tabular}
  \end{center}
\end{table}

%% file: tables/statistics_of_lexicons.tex
\begin{table*}[htp]
  \caption{Statistics of the lexicons.
  The \textit{Single}, \textit{Two}, and \textit{Three} represent the number of characters in the word, respectively.
  The \textit{Other} means the number of words with more characters.
  The \textit{Dimension} is the dimension of the word embedding.}
  \label{tab:lex_statistics}
  \centering
  \begin{center}
  \begin{tabular}{crrrrrc}
    \toprule
	Lexicons & \makebox[1cm][c]{Total} & \makebox[1cm][c]{Single} & \makebox[1cm][c]{Two} & \makebox[1cm][c]{Three} & \makebox[1cm][c]{Other} & Dimension\\
	\midrule
	YJ          &  704.4k &  5.7k &  291.5k &  278.1k &  129.1k & \ \ 50d\\
	LS          & 1292.6k & 18.5k &  347.7k &  415.6k &  511.3k & 300d\\
	TX (v0.1.0) & 8824.3k & 22.7k & 2031.2k & 2034.6k & 4735.8k & 200d\\
    \bottomrule
  \end{tabular}
  \end{center}
\vspace{-0.8em}
\end{table*}

%% file: tables/word_match.tex
\begin{table*}[htp]
  \caption{Match statistics on each lexicon for different datasets. We have done a segmentation process for MSRA and Ontonotes 4.0.}
  \label{tab:word-match}
  \centering
  \begin{center}
  \begin{tabular}{rcccccc}
    \toprule
\multicolumn{1}{r}{Datasets} & \makecell[c]{Char Length\\Max.} & \makecell[c]{Char Length\\Avg.} & \makecell[c]{Matching\\Word Length} & Lex. YJ & Lex. LS & Lex. TX \\
    \midrule
\multirow{2}{*}{Weibo} & \multirow{2}{*}{175} & \multirow{2}{*}{54} & Avg. & \ \ 22 & \ \ 27 & \ \ 43 \\
 &  &  & Max. & 133 & 167 & 164 \\
 \cmidrule(r){4-7}
\multirow{2}{*}{Resume} & \multirow{2}{*}{178} & \multirow{2}{*}{32} & Avg. & \ \ 25 & \ \ 24 & \ \ 40 \\
 &  &  & Max. & 159 & 149 & 215 \\
 \cmidrule(r){4-7}
\multirow{2}{*}{Ontonotes   4.0} & \multirow{2}{*}{273} & \multirow{2}{*}{37} & Avg. & \ \ 19 & \ \ 19 & \ \ 31 \\
 &  &  & Max. & 205 & 185 & 367 \\
 \cmidrule(r){4-7}
\multirow{2}{*}{MSRA} & \multirow{2}{*}{281} & \multirow{2}{*}{46} & Avg. & \ \ 25 & \ \ 25 & \ \ 40\\
 &  &  & Max. & 143 & 166 & 314 \\
    \bottomrule
  \end{tabular}
  \end{center}
\vspace{-0.8em}
\end{table*}

%% file: tables/hyper_parameters_range.tex
\begin{table*}[htp]
  \caption{Range of Hyper-parameters.}
  \label{tab:range_of_hyper-parameters}
  \centering
  \begin{center}
  \begin{tabular}{rp{5cm}c}
    \toprule
Hyper-parameter & Caption & Range \\
    \midrule
warmup & The warm-up ratio. & {[}0.05, 0.1, 0.2, 0.3, 0.4{]} \\
batch\_size & The batch size. & {[}8, 10, 16{]} \\
char\_embed\_dropout & The dropout rate of the character embedding. & {[}0.3,  0.4, 0.5{]} \\
word\_embed\_dropout & The dropout rate of the word embedding. & {[}0.001, 0.002, 0.003{]} \\
lr & The learning rate. & {[}0.001, 0.002, 0.003{]} \\
multi-head & The number of multi-head and the dimensions of each head. & {[}8-16, 8-20, 8-32, 12-16, 12-20{]} \\
fc\_dropout1 & The dropout rate of the linear layer in Transformer and InterFormer. & {[}0, 0.2, 0.4{]} \\
fc\_dropout2 & The dropout rate of the linear layer in NFLAT. & {[}0, 0.2, 0.4{]} \\
attn\_dropout & The dropout rate of the weight for self-attention and inter-attention. & {[}0, 0.1, 0.2{]} \\
is\_less\_head & Whether the number of self-attention heads is half of that of the inter-attention heads. & {[}True, False{]}\\
    \bottomrule
  \end{tabular}
  \end{center}
\vspace{-0.8em}
\end{table*}

%% file: tables/study_case_1.tex
\begin{table*}[htp]
  \caption{The lack of annotations on the test dataset on Ontonotes 4.0.}
  \label{tab:case_1}
  \centering
  \begin{center}
\begin{tabular}{rcccccccccc}
\toprule
\multirow{2}{0.14\textwidth}{\textbf{Sentence \#1}\\ \small{(Truncated)}} & \multicolumn{10}{c}{... 又开始了\textcolor{cyan}{九龙街}的建设 ...} \\
 & \multicolumn{10}{c}{\textit{... Construction of \textcolor{cyan}{Kowloon Street} started again ...}} \\
\cmidrule(r){2-11}
Characters & 又 & 开 & 始 & 了 & \textcolor{cyan}{九} & \textcolor{cyan}{龙} & \textcolor{cyan}{街} & 的 & 建 & 设 \\
Gold Labels & O & O & O & O & \textit{\textcolor{lightgray}{O}} & \textit{\textcolor{lightgray}{O}} & \textit{\textcolor{lightgray}{O}} & O & O & O \\
FLAT & O & O & O & O & B-LOC & M-LOC & E-LOC & O & O & O \\
    \rowcolor{lime!20}
NFLAT & O & O & O & O & B-LOC & M-LOC & E-LOC & O & O & O \\
\midrule
\multirow{2}{0.14\textwidth}{\textbf{Sentence \#2}\\ \small{(Truncated)}} & \multicolumn{10}{c}{... \textcolor{orange}{梁}乡长眉头深锁 ...} \\
 & \multicolumn{10}{c}{\textit{... Village Chief \textcolor{orange}{Liang} frowned deeply ...}} \\
\cmidrule(r){2-11}
Characters &  & ... & ... & \textcolor{orange}{梁} & 乡 & 长 & 眉 & 头 & 深 & 锁 \\
Gold Labels &  & ... & ... & \textit{\textcolor{lightgray}{O}} & O & O & O & O & O & O \\
FLAT &  & ... & ... & B-PER & M-PER & E-PER & O & O & O & O \\
    \rowcolor{lime!20}
NFLAT & ... & ... & ... & S-PER & O & O & O & O & O & O \\
\midrule
\multirow{2}{0.14\textwidth}{\textbf{Sentence \#3}\\ \small{(Truncated)}} & \multicolumn{10}{c}{... 一千两百多位\textcolor{red}{萨国}民众在 ...} \\
 & \multicolumn{10}{c}{\textit{... More than 1,200 people from the \textcolor{red}{Republic of El Salvador} ...}} \\
\cmidrule(r){2-11}
Characters & 一 & 千 & 两 & 百 & 多 & 位 & \textcolor{red}{萨} & \textcolor{red}{国} & 民 & 众 \\

Gold Labels & O & O & O & O & O & O & \textit{\textcolor{lightgray}{O}} & \textit{\textcolor{lightgray}{O}} & O & O \\
FLAT & O & O & O & O & O & O & O & O & O & O \\
    \rowcolor{lime!20}
NFLAT & O & O & O & O & O & O & B-GPE & E-GPE & O & O \\
\bottomrule
\end{tabular}
  \end{center}
\vspace{-0.8em}
\end{table*}

%% file: tables/study_case_2.tex
\begin{sidewaystable*}[htp]
  \caption{Case study comparing FLAT vs NFLAT on Ontonotes 4.0.}
  \label{tab:case_2}
  \centering
  \begin{center}
\begin{tabular}{rccccccccccccc}
\toprule
\multirow{2}{0.08\textwidth}{\textbf{Sentence \#1}\\ \small{(Truncated)}} & \multicolumn{13}{c}{... \textcolor{teal}{美国必纯士国际实业有限公司} ...} \\
 & \multicolumn{13}{c}{\textit{... \textcolor{teal}{American Betwons International Industrial Co., Ltd.} ...}} \\
\cmidrule(r){2-14}
Characters & \textcolor{teal}{美} & \textcolor{teal}{国} & \textcolor{teal}{必} & \textcolor{teal}{纯} & \textcolor{teal}{士} & \textcolor{teal}{国} & \textcolor{teal}{际} & \textcolor{teal}{实} & \textcolor{teal}{业} & \textcolor{teal}{有} & \textcolor{teal}{限} & \textcolor{teal}{公} & \textcolor{teal}{司} \\

Gold Labels & B-ORG & M-ORG & M-ORG & M-ORG & M-ORG & M-ORG & M-ORG & M-ORG & M-ORG & M-ORG & M-ORG & M-ORG & E-ORG \\
FLAT & B-GPE & \textit{\textcolor{lightgray}{E-GPE}} & \textit{\textcolor{lightgray}{O}} & \textit{\textcolor{lightgray}{O}} & \textit{\textcolor{lightgray}{O}} & \textit{\textcolor{lightgray}{B-ORG}} & M-ORG & M-ORG & M-ORG & M-ORG & M-ORG & M-ORG & E-ORG \\
    \rowcolor{lime!20}
NFLAT & B-ORG & M-ORG & M-ORG & M-ORG & M-ORG & M-ORG & M-ORG & M-ORG & M-ORG & M-ORG & M-ORG & M-ORG & E-ORG \\
\midrule
\multirow{2}{0.08\linewidth}{\textbf{Sentence \#2}\\ \small{(Truncated)}} & \multicolumn{13}{c}{... 中部\textcolor{cyan}{秀姑恋山区}的红桧林 ...} \\
 & \multicolumn{13}{c}{\textit{... The red cypress forest in the \textcolor{cyan}{Xiugulian Mountains} in central ...}} \\
\cmidrule(r){2-14}
Characters & 中 & 部 & \textcolor{cyan}{秀} & \textcolor{cyan}{姑} & \textcolor{cyan}{恋} & \textcolor{cyan}{山} & \textcolor{cyan}{区} & 的 & 红 & 桧 & 林 & ... \\
Gold Labels & O & O & B-LOC & M-LOC & M-LOC & M-LOC & E-LOC & O & O & O & O & ... \\
FLAT & \textit{\textcolor{lightgray}{B-LOC}} & \textit{\textcolor{lightgray}{M-LOC}} & \textit{\textcolor{lightgray}{M-LOC}} & M-LOC & M-LOC & M-LOC & E-LOC & O & \textit{\textcolor{lightgray}{B-LOC}} & \textit{\textcolor{lightgray}{M-LOC}} & \textit{\textcolor{lightgray}{E-LOC}} & ... \\
    \rowcolor{lime!20}
NFLAT & O & O & B-LOC & M-LOC & M-LOC & M-LOC & E-LOC & O & O & O & O & ... & \\
\midrule
\multirow{2}{0.08\linewidth}{\textbf{Sentence \#3}\\ \small{(Truncated)}} & \multicolumn{13}{c}{... \textcolor{teal}{龙山国中}训导主任\textcolor{orange}{陈采卿} ...} \\
 & \multicolumn{13}{c}{\textit{... \textcolor{orange}{Caiqing Chen}, the director of \textcolor{teal}{Longshan junior high school} ...}} \\
\cmidrule(r){2-14}
Characters & \textcolor{teal}{龙} & \textcolor{teal}{山} & \textcolor{teal}{国} & \textcolor{teal}{中} & 训 & 导 & 主 & 任 & \textcolor{orange}{陈} & \textcolor{orange}{采} & \textcolor{orange}{卿} & ... \\
Gold Labels & B-ORG & M-ORG & M-ORG & E-ORG & O & O & O & O & B-PER & M-PER & E-PER & ... \\
FLAT & B-ORG & M-ORG & M-ORG & \textit{\textcolor{lightgray}{M-ORG}} & \textit{\textcolor{lightgray}{M-ORG}} & \textit{\textcolor{lightgray}{E-ORG}} & O & O & B-PER & M-PER & E-PER & ... \\
    \rowcolor{lime!20}
NFLAT & B-ORG & M-ORG & M-ORG & E-ORG & O & O & O & O & B-PER & M-PER & E-PER & ... & \\
\midrule
\multirow{2}{0.08\linewidth}{\textbf{Sentence \#4}\\ \small{(Truncated)}} & \multicolumn{13}{c}{... \textcolor{red}{津巴布韦}选手\textcolor{orange}{斯·埃万} ...} \\
 & \multicolumn{13}{c}{\textit{... The competitor from \textcolor{red}{Zimbabwe}, \textcolor{orange}{Si Aiwan} ...}} \\
\cmidrule(r){2-14}
Characters & \textcolor{red}{津} & \textcolor{red}{巴} & \textcolor{red}{布} & \textcolor{red}{韦} & 选 & 手 & \textcolor{orange}{斯} & \textcolor{orange}{·} & \textcolor{orange}{埃} & \textcolor{orange}{万} & ... \\
Gold Labels & B-GPE & M-GPE & M-GPE & E-GPE & O & O & B-PER & M-PER & M-PER & E-PER & ... \\
FLAT & \textit{\textcolor{lightgray}{O}} & \textit{\textcolor{lightgray}{O}} & \textit{\textcolor{lightgray}{O}} & \textit{\textcolor{lightgray}{O}} & O & O & \textit{\textcolor{lightgray}{O}} & \textit{\textcolor{lightgray}{O}} & \textit{\textcolor{lightgray}{O}} & \textit{\textcolor{lightgray}{O}} & ... \\
    \rowcolor{lime!20}
NFLAT & B-GPE & M-GPE & M-GPE & E-GPE & O & O & B-PER & M-PER & M-PER & E-PER & ... & & \\
\bottomrule
\end{tabular}
  \end{center}
\vspace{-0.8em}
\end{sidewaystable*}